\documentclass[a4paper, 10pt, conference]{ieeeconf}      
\usepackage{FG2021}

\usepackage{soul}
\usepackage{url}
\usepackage[hidelinks]{hyperref}
\usepackage[utf8]{inputenc}

\usepackage{graphicx}
\usepackage{epsfig}
\usepackage{amsmath}
\usepackage{booktabs}
\usepackage{algorithm}
\usepackage{algorithmic}
\usepackage{amssymb}
\usepackage{color}
\usepackage{hhline}
\usepackage{stfloats}

\usepackage[accsupp]{axessibility} 

\definecolor{Gray}{gray}{0.85}

\newcommand{\eqnref}[1]{Equation~(\ref{eqn:#1})}
\newcommand{\figref}[1]{Fig.~\ref{fig:#1}}
\newcommand{\tabref}[1]{Table~\ref{tbl:#1}}
\newcommand{\secref}[1]{Section~\ref{sec:#1}}

\newcommand{\NetPipeline}{
\begin{figure}[t]
    \centering
    \includegraphics[width=1\linewidth]{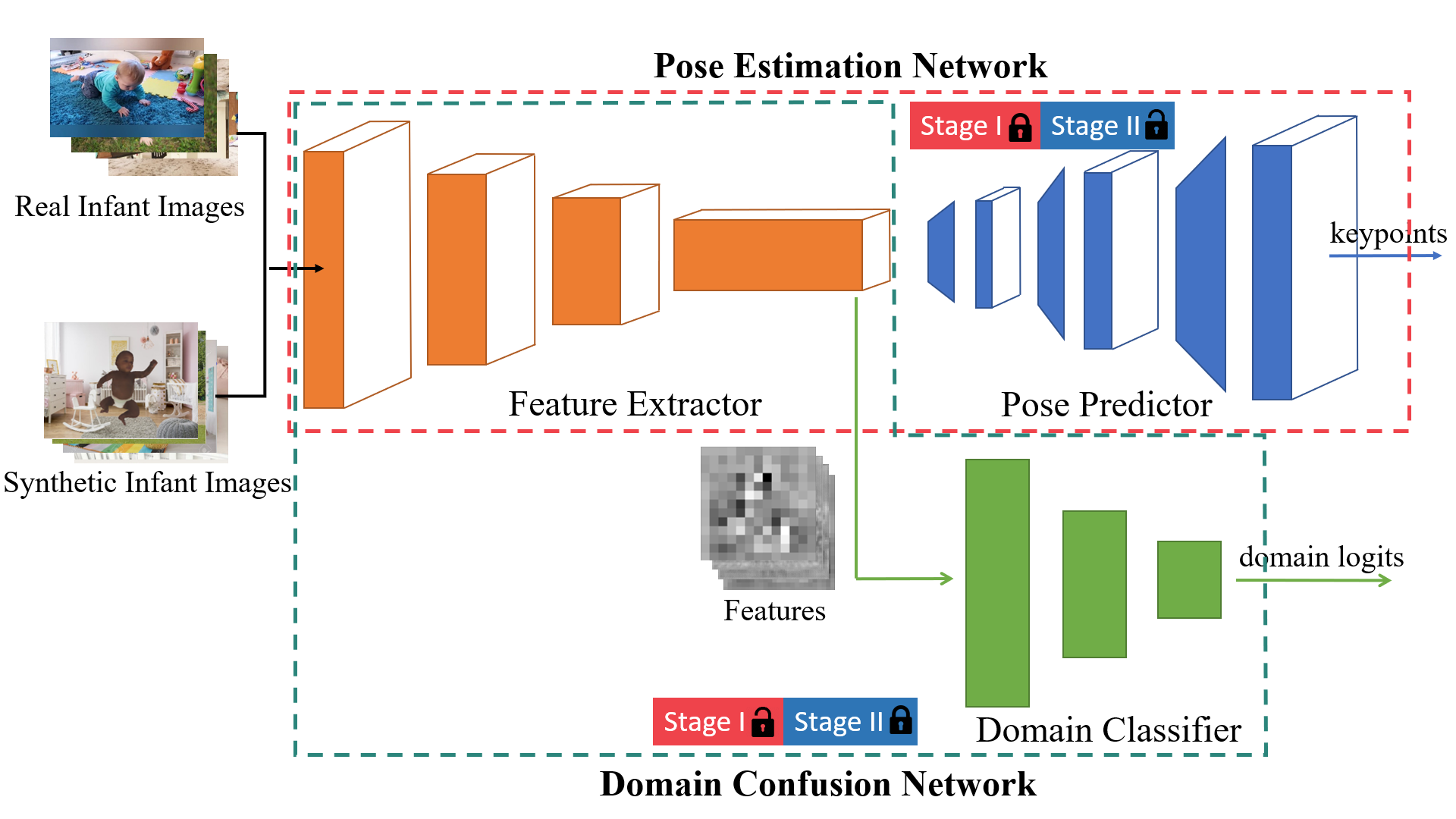}
    \caption{An overview architecture of our fine-tuned domain-adapted infant pose (FiDIP) framework, composed of two sub-networks: pose estimation network (red-dot box) and domain confusion network (blue-dot box). Main components of FiDIP include a feature extractor (orange), a pose predictor (blue), and a domain classifier (green).} 
    \label{fig:NetPipeline}
            \vspace{-.15in}
\end{figure}
}

\newcommand{\visualized}{
\begin{figure}[t]
    \centering
    \includegraphics[width=1\linewidth]{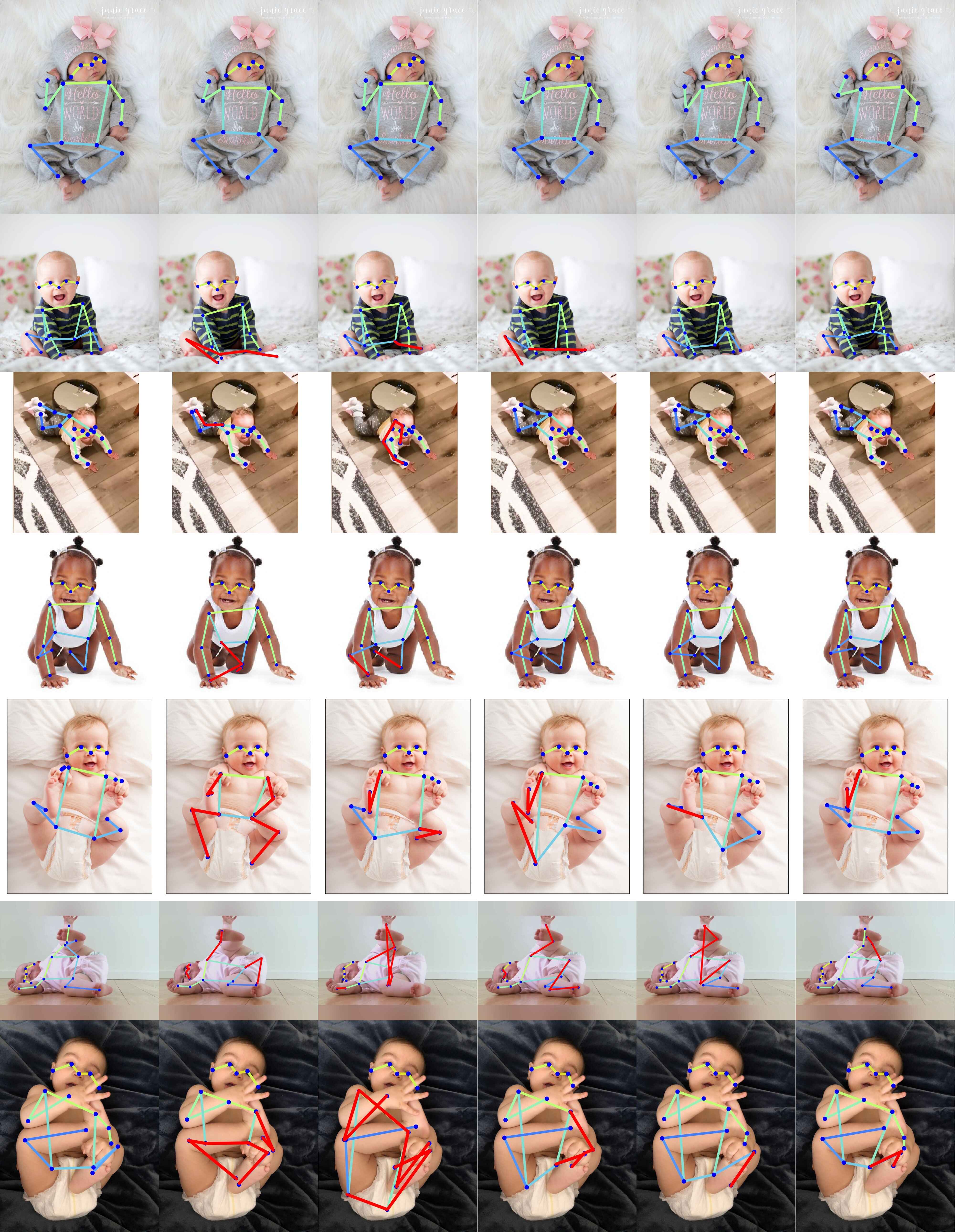}
    \caption{Samples of infant pose prediction results of DarkPose: AP=65.9 (2nd column), FasterR-CNN: AP=70.1 (3rd column), SimpleBaseline: AP=82.4 (4rd column), DarkPose: AP=88.5 (5th column), and our SimpleBaseline+FiDIP: AP=91.1 (6th column) on SyRIP Test100, which are listed in \tabref{comp}. The 1st column is the visualization of groundtruth. Incorrect predictions are highlighted in red. Note: more visualized results are given in \textit{Supplementary Materials}.} 
    \label{fig:visualized}
            \vspace{-.15in}
\end{figure}
}

\newcommand{\Distri}{
\begin{figure}[t]
\centering
\includegraphics[width=1\linewidth]{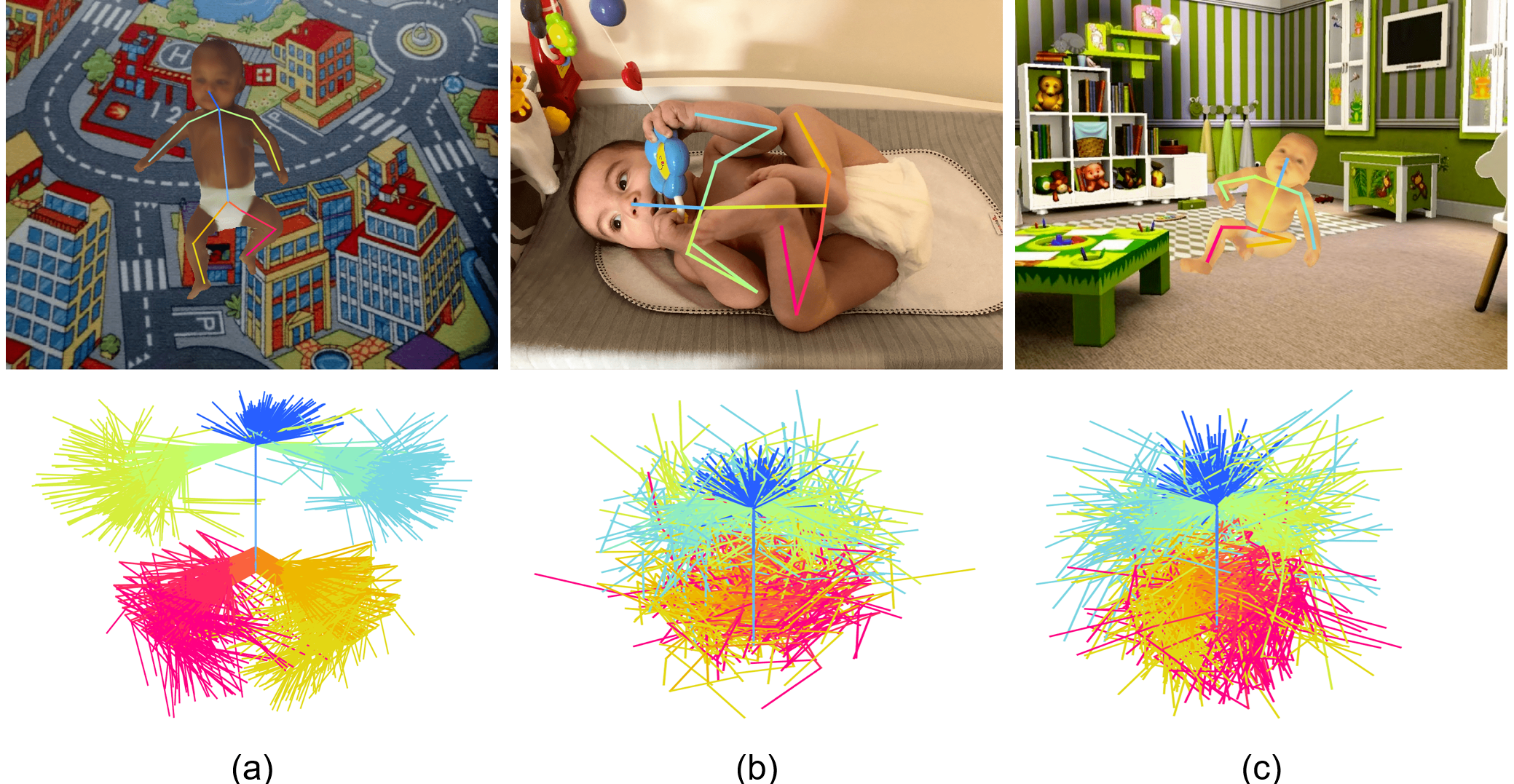}
\caption{A sample image and pose distribution of (a)  MINI-RGBD dataset, (b) the real part of the SyRIP dataset, and  (c)  the synthetic part of the SyRIP dataset. The first row shows a sample image from each dataset with its groundtruth labels. The second row shows the pose distribution of 200 images that are randomly selected from each dataset, in which colors of different body parts correspond to the colors of body parts of figures in the first row.  For the pose distribution, we normalize all images based on the infant bounding box to scale them into similar sizes, then align them based on their torso with upward head. To better represent the poses, we also ignore the points for ears and eyes when  visualizing the joints.}
\label{fig:Distri}
     \vspace{-.15in}
\end{figure}
}

\newcommand{\SynPipeline}{
\begin{figure}[t]
    \centering
    \includegraphics[width=1.0\linewidth]{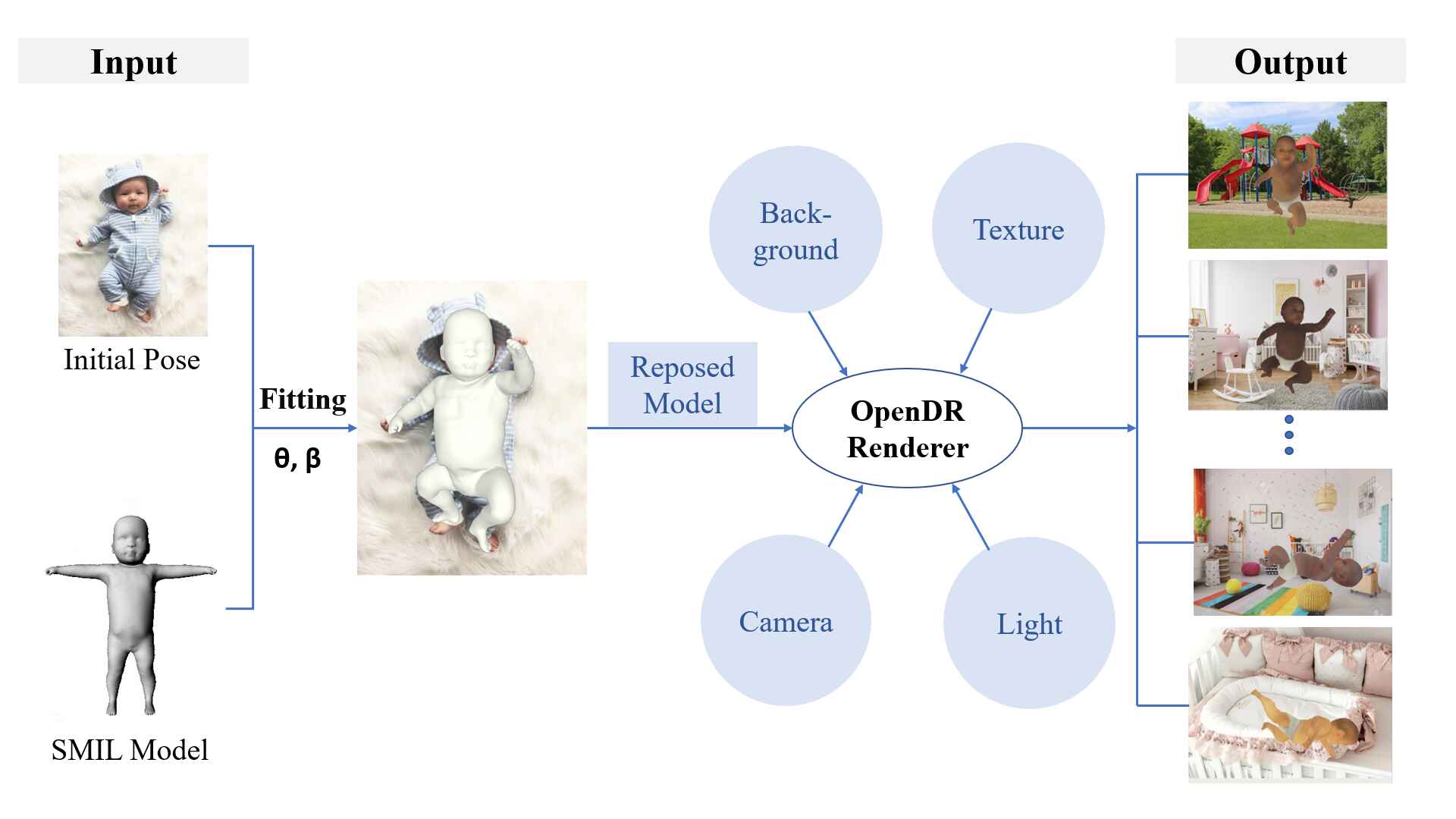}
    \caption{Our pipeline of synthetic infant image generation. 3D infant body models are posed by fitting SMIL model pose and shape parameters into real infant images. The generated model is reposed by adding variances to pose coefficients, $\theta$. Then output images are rendered using random background images, texture maps on the body, lighting, and camera positions.} 
    \label{fig:SynPipeline}
     \vspace{-.15in}
\end{figure}
}

\newcommand{\fDistri}{
\begin{figure}[thpb]
\centering
\includegraphics[width=1\linewidth]{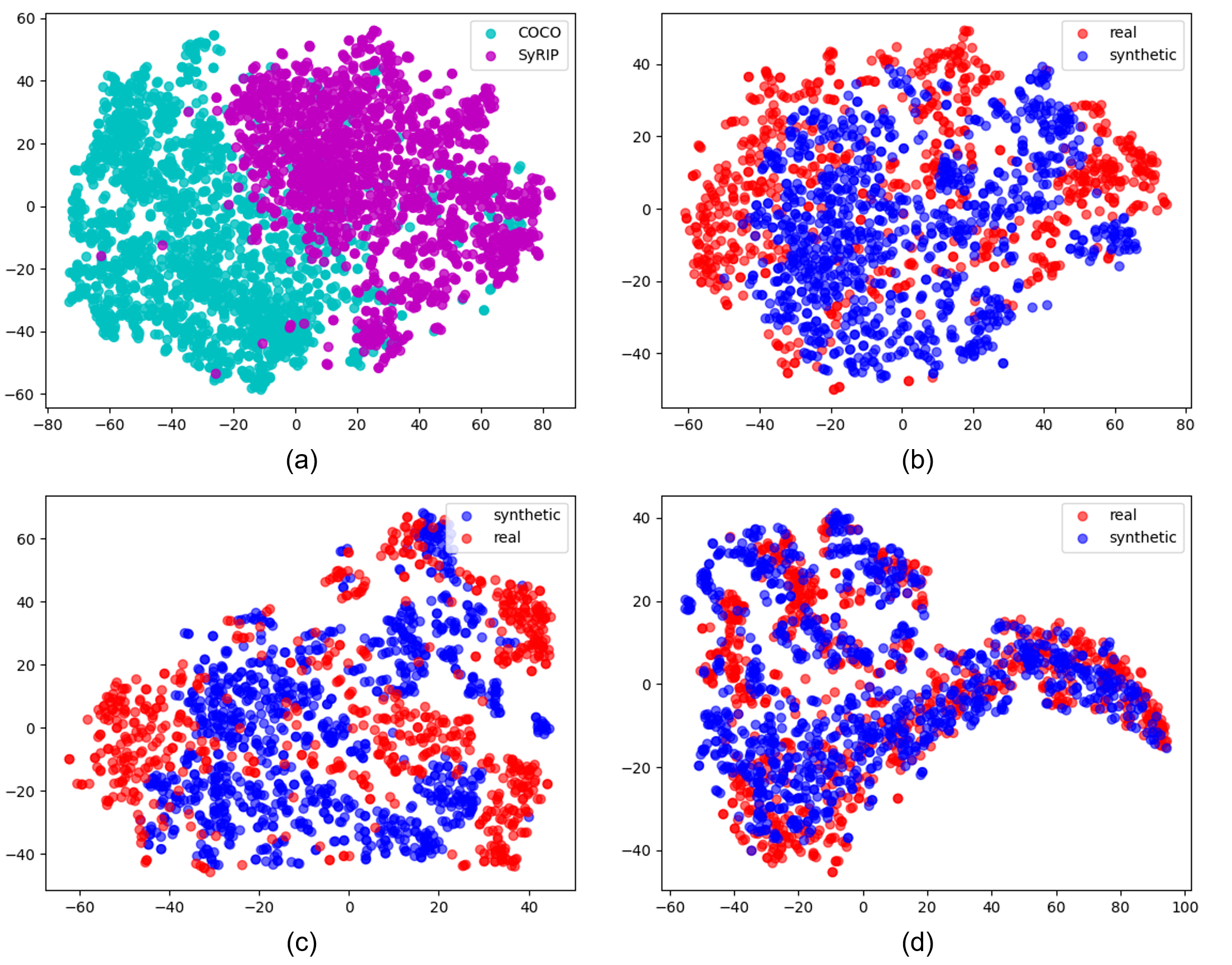}
\caption{t-SNE visualized features of (a) 700 random samples of COCO Val2017 vs. 700 real images of SyRIP, and 1000 synthetic images vs. 700 real images of SyRIP dataset extracted by (b) original  SimpleBaseline-50, (c) method \textbf{i} (fine-tuning without domain adaptation), and (d) method \textbf{n} (fine-tuning with domain adaptation).}
    \vspace{-.15in}
\label{fig:fDistri}
    \vspace{-.15in}
\end{figure}
}

\newcommand{\visualizedEXT}{
\begin{figure*}[ht]
    \centering
    \includegraphics[width=1\linewidth]{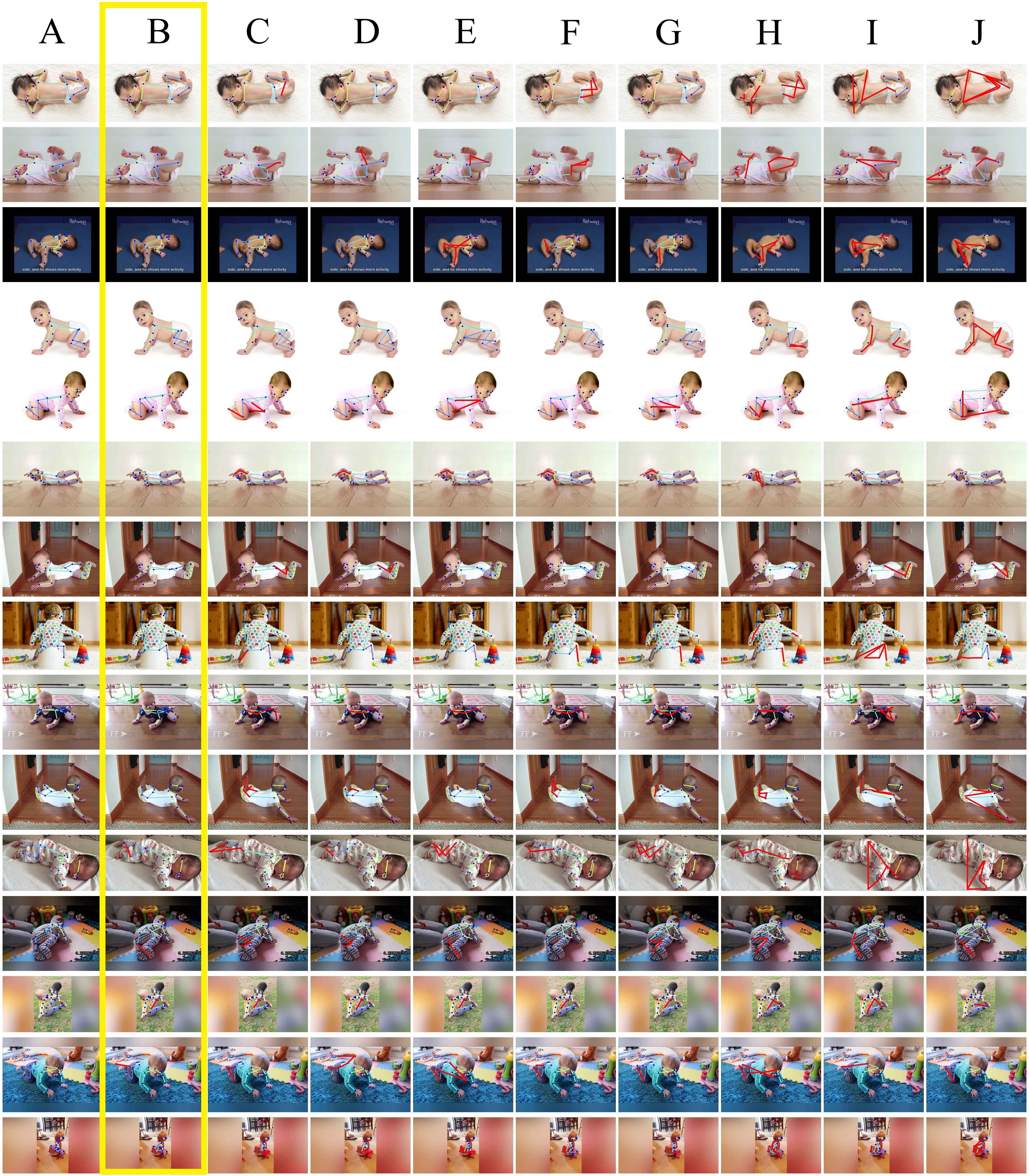}
    \caption{Samples of infant pose prediction results of SOTA models on SyRIP Test500, which are listed in \tabref{comp}. The column A is the visualization of groundtruth poses, the column B in yellow box is our SimpleBaseline+FiDIP model results. The following columns show the results of models ordered based on the in AP accuracy when tested on SyRIP Test500. The  columns C to J are: DarkPose:AP=98.0, DarkPose:AP=97.7, SimpleBaseline:AP=97.6, DarkPose:AP=97.4, SimpleBaseline:AP=97.3, DarkPose:AP=95.2, FasterR-CNN:AP=93.4, and FasterR-CNN:AP=91.9.  Incorrect predictions are highlighted in red in each image.} 
    \label{fig:visualizedEXT}
            \vspace{-.15in}
\end{figure*}
}

\newcommand{\ComplexPose}{
\begin{figure}[H]
    \centering
    \includegraphics[width=0.8\linewidth]{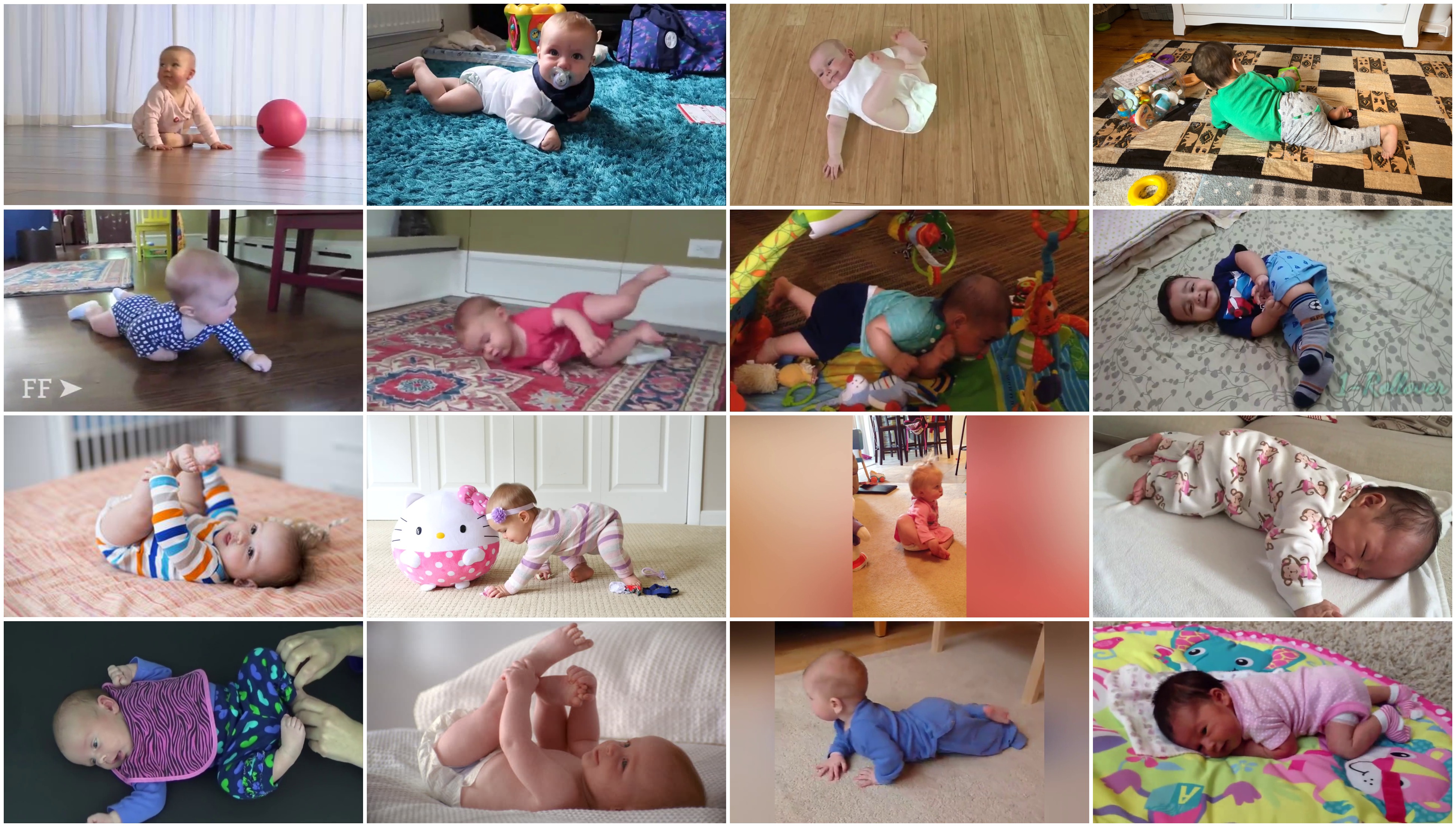}
    \caption{Complex poses in SyRIP Test100.} 
    \label{fig:ComplexPose}
     \vspace{-.15in}
\end{figure}
}
\usepackage{amssymb}
\usepackage{multicol}
\usepackage{multirow}
\usepackage{threeparttable}

\newcommand{\syripabl}{
\begin{table*}[t]
\caption{\footnotesize Performance comparison of three SOTA pose estimation models (SimpleBaseline, DarkPose, Pose-MobileNet) fine-tuned on MINI-RGBD, SyRIP-syn (synthesized data only) and SyRIP whole set and tested on SyRIP Test100.} 
\scriptsize
\begin{center}
\begin{tabular}{l||ccc||cccccc}
Train Set  & Method  & Backbone  & Input size  & AP   & AP50 & AP75 & AR   & AR50 & AR75 \\ \hline \hline
MINI-RGBD  & \multicolumn{1}{c|}{}  & \multicolumn{1}{c|}{}   &  & 69.2 & 95.8 & 78.0 & 72.4 & 97.0 & 81.0 \\ 
\cline{1-1} \cline{5-10} 
SyRIP-syn & \multicolumn{1}{c|}{\multirow{2}*{SimpleBaseline}}  & \multicolumn{1}{c|}{\multirow{2}*{ResNet-50}}  & \multirow{2}*{384x288}                 & 85.3 & 97.1 & 91.8 & 87.4 & 98.0 & 93.0 \\ 
\cline{1-1} \cline{5-10} 
SyRIP    & \multicolumn{1}{c|}{}  & \multicolumn{1}{c|}{}  &       & \textbf{90.1} & 98.5 & 97.2 & 91.6 & 99.0 & 98.0 \\ 
\hline \hline
MINI-RGBD                  & \multicolumn{1}{c|}{}   & \multicolumn{1}{c|}{} &  & 85.2 & 98.6 & 95.3 & 87.0 & 99.0 & 96..0 \\ 
\cline{1-1} \cline{5-10} 
SyRIP-syn       & \multicolumn{1}{c|}{\multirow{2}*{DarKPose}}
& \multicolumn{1}{c|}{\multirow{2}*{HRNet-W48}}                             & \multirow{2}*{384x288} 
& 91.4 & 98.5 & 98.5 & 92.7 & 99.0 & 99.0 \\ 
\cline{1-1} \cline{5-10} 
SyRIP          & \multicolumn{1}{c|}{}           & \multicolumn{1}{c|}{}     &          & \textbf{92.7} & 98.5 & 98.5 & 93.9 & 99.0 & 99.0     \\ \hline \hline
MINI-RGBD            & \multicolumn{1}{c|}{} & \multicolumn{1}{c|}{} &   & 12.3 & 38.1 & 3.8 & 21.6 & 52.0 & 14.0 \\
\cline{1-1} \cline{5-10} 
SyRIP-syn  & \multicolumn{1}{c|}{\multirow{2}*{Pose-MobileNet}} 
& \multicolumn{1}{c|}{\multirow{2}*{MobileNetV2}}                             & \multirow{2}*{224x224}                    & 60.3 & 91.1 & 62.7 & 68.4 & 95.0 & 72.0 \\ \cline{1-1} \cline{5-10} 
SyRIP    & \multicolumn{1}{c|}{}                             & \multicolumn{1}{c|}{}                         &                                    & \textbf{78.9} & 97.2 & 90.6 & 84.2 & 98.0 & 94.0        \\ \hline
\multicolumn{10}{l}{\scriptsize Best AP for each method is highlighted in bold font.}
\end{tabular}
\label{tbl:syripabl}
\end{center}
\vspace{-.2in}
\end{table*}
}

\newcommand{\comp}{
\begin{table*}[t]
\caption{\footnotesize Performance comparison between SimpleBaseline model applied FiDIP method and the SOTA pose estimators on the COCO Val2017 and SyRIP test datasets.}
\scriptsize
\begin{center}
 \begin{tabular}{l || c  c  c  c  c  c  c c c c}
\textbf{Pose Estimation}  & \textbf{Backbone}  & \textbf{Input }  & \textbf{COCO\_Val2017}       &\textbf{SyRIP\_Test500}  &\multicolumn{6}{c}{\textbf{SyRIP\_Test100}}\\
\cline{6-11}
 \textbf{Model} & \textbf{Network} & \textbf{Image Size}  &\textbf{AP}   &\textbf{AP} &\textbf{AP}   &\textbf{AP50}  &\textbf{AP75} &\textbf{AR}   &\textbf{AR50}  &\textbf{AR75} \\
\hline
\hline
Faster R-CNN \cite{wu2019detectron2}       & ResNet-50-FPN  & Flexible      & 65.5    & 93.4     & 70.1  & 97.7  & 73.8  & -  & -  & - \\ \hline
Faster R-CNN \cite{wu2019detectron2}       & ResNet-101-FPN  & Flexible     & 66.1    & 91.9     & 64.4  & 95.2  & 71.5   & -  & -  & - \\ \hline
DarkPose \cite{zhang2020distribution}      & ResNet-50  & 128$\times$96     & 62.6    & 95.2     & 65.9  & 94.8  & 66.7  & 69.2  & 96.0 & 71.0  \\ \hline
DarkPose \cite{zhang2020distribution}      & HRNet-W48  & 128$\times$96     & 71.9    & 97.4     & 82.1  & 98.6  & 92.2  & 83.6  & 99.0 & 93.0  \\ \hline
DarkPose \cite{zhang2020distribution}      & HRNet-W32  & 256$\times$192    & 75.6    & 97.7     & 88.5  & 98.4  & 98.4  & 90.1  & 99.0 & 99.0 \\ \hline
DarkPose \cite{zhang2020distribution}      & HRNet-W48  & 384$\times$288    & \textbf{76.9}    & 98.0     & 88.5  & 98.5  & 98.5  & 90.0  & 99.0 & 99.0   \\ \hline
SimpleBaseline \cite{xiao2018simple}       & ResNet-50  & 256$\times$192    & 70.4    & 97.3    & 80.4  & 98.5 &92.2  &82.5  & 99.0 & 94.0  \\ \hline
SimpleBaseline \cite{xiao2018simple}       & ResNet-50  & 384$\times$288    & 72.2    & 97.6     & 82.4  & 98.9  & 92.2  & 83.8 & 99.0 & 93.0 \\ \hline
RMPE \cite{fang2017rmpe}                   & VGG\_SSD    & 500$\times$500   & 61.8    & 76.2     & 76.3  & 82.4  & 78.3 & -  & -  & - \\\hline
UDP \cite{Huang_2020_CVPR}                 & HRNet-W32  & 256$\times$192    & 75.2    & 81.2     & 79.8  & 86.2  & 88.4 &71.3  &73.2 &74.0  \\ \hline
UDP \cite{Huang_2020_CVPR}                 & ResNet-50  & 256$\times$192    & 71.7    & 83.4     & 78.2  & 80.2 &77.4  &75.1 &75.4 &76.7    \\ \hline
UDP \cite{Huang_2020_CVPR}                 & ResNet-152  & 384$\times$288   & 74.7    & 84.2     & 79.1  & 81.5 & 82.8 & 81.1& 80.4&79..6   \\ \hline
\textbf{SimpleBaseline + FiDIP (Ours)}     & ResNet-50  & 384$\times$288    & 59.6    & \textbf{98.3} & \textbf{91.1}& 98.5 & 98.5 & 92.6 & 99.0 & 99.0 \\ \hline

\multicolumn{11}{l}{\scriptsize Best results for different test sets are highlighted in bold fonts.}
\end{tabular}
\label{tbl:comp}
\end{center}
\vspace{-.2in}
\end{table*}
}

\newcommand{\ablation}{
\begin{table}[t]
\caption{\footnotesize Ablation study of FiDIP  on SyRIP Test100 dataset with resolution of 384$\times$288. DC stands for domain classifier. SimpleBaseline-50 \cite{xiao2018simple} is provided here for a baseline comparison. } 
\vspace{-.15in}
\scriptsize
\begin{center}
\begin{threeparttable}
 \begin{tabular}{c || c   c   c  c  c}
\textbf{Method}  & \textbf{Training} & \textbf{Domain }    & \textbf{Pre-train}  &  \textbf{Update}   &\textbf{SyRIP}  \\ 
\textbf{}  & \textbf{Data} & \textbf{Adaptation}    & \textbf{DC}  &  \textbf{Layers}   &\textbf{Test-AP}  \\ 
\hline
\hline

SB-50(*) & - &- & - &-  & 82.4 \\ 
\hline 
\hline 
\textbf{a}	& 1000 Syn	&  $\times$      & -	 & Res 4, 5 &    84.1  \\ \hline
\textbf{b}	& 1000 Syn	&  $\times$	     & -	 & Res 5  &    85.3    \\ \hline
\textbf{c}	& 1000 Syn	&  $\checkmark$	 & $\times$ & Res 4, 5 &  84.6       \\ \hline
\textbf{d}	& 1000 Syn	&  $\checkmark$	 &  $\checkmark$ & Res 4, 5& 85.3 \\ \hline
\textbf{e}	& 1000 Syn	&  $\checkmark$	 & $\times$  & Res 5 &  86.3\\ \hline
\textbf{f}	& 1000 Syn	&  $\checkmark$	 &  $\checkmark$	& Res 5 & 85.5  \\ \hline

\textbf{g}	& 200  Real	&  $\times$      & -	 & Res 4, 5 &    87.1  \\ \hline
\textbf{h}	& 200  Real	&  $\times$    	 & -	 & Res 5  &    86.9    \\ \hline

\textbf{i}	& 1200 R+S	&  $\times$	     & -	   & Res 4, 5 & 90.1 \\ \hline
\textbf{j}	& 1200 R+S	&  $\times$	     & -	   & Res 5  &  90.0\\ \hline
\textbf{k}	& 1200 R+S	&  $\checkmark$	 & $\times$	& Res 5 &  90.2\\ \hline
\textbf{l}	& 1200 R+S	&  $\checkmark$	 &  $\checkmark$	& Res 5 & 90.3\\ \hline
\textbf{m}	& 1200 R+S	&  $\checkmark$	 & $\times$   & Res 4, 5 &  90.3\\ \hline
\textbf{n}	& 1200 R+S	&  $\checkmark$	 & $\checkmark$  & Res 4, 5 & \textbf{91.1}\\  \hline
\end{tabular}
\begin{tablenotes}
    \item[*] SimpleBaseline-50
  \end{tablenotes}
\end{threeparttable}
\label{tbl:ablation}
\end{center}
\vspace{-.2in}
\end{table}
}

\newcommand{\mobile}{
\begin{table*}[t]
\caption{\footnotesize Evaluating the generality of our FiDIP method to different SOTA models on the SyRIP Test100. } 
\scriptsize
\begin{center}
\begin{tabular}{l||cccc||cccccc}
Method  & Backbone  & Input size  & \# Params   & GFLOPs    & AP   & AP50 & AP75 & AR   & AR50 & AR75 \\ 
\hline \hline
SimpleBaseline            & \multicolumn{1}{c|}{}    & \multicolumn{1}{c|}{}    & \multicolumn{1}{c|}{} &   & 82.4 & 98.9 & 92.2 & 83.8 & 99.0 & 93.0 \\ \cline{1-1} \cline{6-11} 
SimpleBaseline + Finetune & \multicolumn{1}{c|}{\multirow{1}*{ResNet-50}}  & \multicolumn{1}{c|}{\multirow{1}*{384x288}}      & \multicolumn{1}{c|}{\multirow{1}*{32.42M}}                        & \multirow{1}*{20.23}                       & 90.1 & 98.5 & 97.2 & 91.6 & 99.0 & 98.0 \\ \cline{1-1} \cline{6-11} 
SimpleBaseline + FiDIP    & \multicolumn{1}{|c|}{}     & \multicolumn{1}{c|}{}      & \multicolumn{1}{c|}{}      &                        & \textbf{91.1} & 98.5 & 98.5 & 92.6 & 99.0 & 99.0 \\ 
\hline \hline
DarkPose                  & \multicolumn{1}{c|}{}   & \multicolumn{1}{c|}{} & \multicolumn{1}{c|}{} &  & 88.5 & 98.5 & 98.5 & 90.0 & 99.0 & 99.0 \\ \cline{1-1} \cline{6-11} 
DarkPose + Finetune       & \multicolumn{1}{c|}{\multirow{1}*{HRNet-W48}}                             & \multicolumn{1}{c|}{\multirow{1}*{384x288}}                         & \multicolumn{1}{c|}{\multirow{1}*{60.65M}}                        & \multirow{1}*{32.88}       & 92.7 & 98.5 & 98.5 & 93.9 & 99.0 & 99.0 \\ \cline{1-1} \cline{6-11} 
DarkPose + FiDIP          & \multicolumn{1}{c|}{}           & \multicolumn{1}{c|}{}     & \multicolumn{1}{c|}{}      &                        & \textbf{93.6} & 98.5 & 98.5 & 94.6 & 99.0 & 99.0 \\ \hline \hline
Pose-MobileNet            & \multicolumn{1}{c|}{} & \multicolumn{1}{c|}{} & \multicolumn{1}{c|}{}  &   & 46.5 & 85.7 & 45.6 & 56.2 & 89.0 & 59.0 \\ \cline{1-1} \cline{6-11} 
Pose-MobileNet + Finetune & \multicolumn{1}{c|}{\multirow{1}*{MobileNetV2}}                             & \multicolumn{1}{c|}{\multirow{1}*{224x224}}                         & \multicolumn{1}{c|}{\multirow{1}*{3.91M}}                        &  \multirow{1}*{0.46}                     & 78.9 & 97.2 & 90.6 & 84.2 & 98.0 & 94.0 \\ \cline{1-1} \cline{6-11} 
Pose-MobileNet + FiDIP    & \multicolumn{1}{c|}{}                             & \multicolumn{1}{c|}{}                         & \multicolumn{1}{c|}{}                        &                        & \textbf{79.3} & 99.0 & 89.4 & 84.1 & 99.0 & 92.0 \\ \hline
\multicolumn{11}{l}{\scriptsize Best AP for each method is highlighted in bold font.}
\end{tabular}
\label{tbl:mobile}
\end{center}
\vspace{-.25in}
\end{table*}
}

\newcommand{\supp}{
\newpage
\onecolumn
\newcommand{\beginsupplement}{%
        \setcounter{table}{0}
        \setcounter{equation}{0}
        \renewcommand{\theequation}{S\arabic{equation}}
        \setcounter{section}{0}
        \renewcommand{\thetable}{S\arabic{table}}%
        \setcounter{figure}{0}
        \renewcommand{\thefigure}{S\arabic{figure}}%
}

\newcommand{\independent}{\protect\mathpalette{\protect\independenT}{\perp}}
\def\independenT##1##2{\mathrel{\rlap{$##1##2$}\mkern2mu{##1##2}}}
\renewcommand\thesection{\Alph{section}}
\beginsupplement
\newcommand{\MYhref}[3][blue]{\href{##2}{\color{##1}{##3}}}

\section{Supplementary Materials}

\subsection{Complex Poses}
Most of the infant poses are very different from those of adults. Especially because of the baby's softer body, the folded poses and occluded joints are more difficult to be recognized or predicted. Some of these typical poses selected from our SyRIP Test100 (complex poses collection) are shown in \figref{ComplexPose}.

\ComplexPose

\subsection{Qualitative  Results} 
In \figref{visualizedEXT}, we exhibit more visualized results for our SimpleBaseline+FiDIP model compared with the other well-performed pose estimation models (where their AP is higher than 90.0 on SyRIP Test500) listed in \tabref{comp}. 

\visualizedEXT
}

\FGfinalcopy 

\IEEEoverridecommandlockouts                              
\def\FGPaperID{101} 

\title{\LARGE \bf Invariant Representation Learning \\for Infant Pose Estimation with Small Data}

\author{\parbox{16cm}{\centering
    {\large Xiaofei Huang, Nihang Fu, Shuangjun Liu, Sarah Ostadabbas\\
    {\normalsize
    Augmented Cognition Lab, Electrical and Computer Engineering Department, \\Northeastern University, Boston, MA, USA\\}}
}}

\begin{document}





\IEEEoverridecommandlockouts\pubid{\makebox[\columnwidth]{978-1-6654-3176-7/21/\$31.00~\copyright{}2021 IEEE \hfill}
\hspace{\columnsep}\makebox[\columnwidth]{ }}

\ifFGfinal
\thispagestyle{empty}
\pagestyle{empty}
\else
\author{Anonymous FG2021 submission\\ Paper ID \FGPaperID \\}
\pagestyle{plain}
\fi
\maketitle

\begin{abstract}
 Infant motion analysis is a topic with critical importance in early childhood development studies. However, while the applications of human pose estimation have become more and more broad, models trained on large-scale adult pose datasets are barely successful in estimating infant poses due to the significant differences in their body ratio and the versatility of their poses. Moreover, the privacy and security considerations hinder the availability of adequate infant pose data required for training of a robust model from scratch. To address this problem, this paper presents (1) building and publicly releasing a hybrid synthetic and real infant pose (SyRIP) dataset with small yet diverse real infant images as well as generated synthetic infant poses and (2) a multi-stage invariant representation learning strategy that could transfer the knowledge from the adjacent domains of adult poses and synthetic infant images into our fine-tuned domain-adapted infant pose (FiDIP) estimation model. In our ablation study, with identical network structure, models trained on SyRIP dataset show noticeable improvement over the ones trained on the only other public infant pose datasets. Integrated with pose estimation backbone networks with varying complexity,  FiDIP performs consistently better than the fine-tuned versions of those models.  One of our best infant pose estimation performers on the state-of-the-art DarkPose model shows mean average precision (mAP) of 93.6\footnote{The code is available at:  \href{https://github.com/ostadabbas/Infant-Pose-Estimation}{github.com/ostadabbas/Infant-PoseEstimation}. The SyRIP dataset can be downloaded at:  \href{https://coe.northeastern.edu/Research/AClab/SyRIP/}{Synthetic and Real Infant Pose (SyRIP)}.}.
\end{abstract}

\section{Introduction}
\label{sec:intro}
Current efforts in machine learning, especially with the recent waves of deep learning models introduced in the last decade, have obliterated records for regression and classification tasks that have previously seen only incremental accuracy improvements. However, this performance comes at a large data cost. There are many other applications that would significantly benefit from the deep learning, where data collection or labeling is expensive and limited. In these domains, which we refer to as ``Small Data'' domains, the challenge we facing is how to learn efficiently with the same performance with less data. One example of these applications with the small data challenges is the problem of infant pose estimation. In infants, long-term monitoring of their poses provide information about their health condition and accurate recognition of these poses can lead to a better early developmental risk assessment and diagnosis \cite{prechtl1990qualitative,hadders1997assessment}. Both motor delays and atypical movements are presented in children with cerebral palsy and are risk indicators for autism spectrum disorders \cite{zwaigenbaum2013early,vyas2019recognition}.

However, current publicly available human pose datasets are predominantly from scenes such as sports, TV shows, and other daily activities performed by adult humans, and none of these datasets provides any specific infants or young children pose images. Beside privacy issues which hamper large-scale data collection from infant, infant pose images differ from available adult pose datasets due to the notable differences in their pose distribution (see \figref{fDistri}(a)) compared to the common adult poses collected from surveillance viewpoints \cite{liu2017vision}. 
These differences mainly come from  (1) shorter limbs and completely different bone to muscle ratio compared to adults; (2) different activities, appearances, and environmental contexts, which together result in sub-optimal performance of the pre-trained models trained on adult poses when tested on infant images  (see \secref{experiments}) with either over-prediction or under-prediction of the limb sizes. 

\NetPipeline

In this paper, towards building a robust infant pose estimation model, we propose a strategy to transfer the pose learning of the existing adult pose estimation models into the infant poses. 
It includes a hybrid synthetic and real infant pose dataset built based on our cross domain inspired augmentation (CDIA) approach and 
a fine-tuned domain-adapted infant pose (FiDIP) estimation network as shown in \figref{NetPipeline}, which is a data-efficient inference model bootstrapped  on both transfer learning and synthetic data augmentation approaches. 
In this paper,  we address the critical small data problem of infant pose estimation by making the following contributions: 
\begin{itemize}
    \item Building and publicly releasing a novel full-annotated hybrid synthetic and real infant pose (SyRIP) dataset via a proposed cross domain inspiration augmentation technique. SyRIP includes a diverse set of real and synthetic infant images, which benefits from (1)  appearance and pose of real infants in images scrapped form web, and (2) the augmented variations in view points, poses, backgrounds, and appearances by synthesizing infant avatars. SyRIP provides advantages over the existing (and very limited) infant pose datasets, by clearly improving the performance of the models trained on it.  

    \item Proposing a fine-tuned domain-adapted infant pose (FiDIP) framework built upon a two-stage training paradigm. In the stage I of training, we fine-tune a pre-trained synthetic/real domain confusion network in a pose-unsupervised manner. In the stage II, we fine-tune a pre-trained pose estimation model under the guidance of stage I trained domain confusion network. Both  networks are updated separately in iterative way. 
   
    \item Achieving two invariant representation learning goals simultaneously. In the FiDIP network, there exist two transfer learning tasks: (1) from adult pose domain into the infant pose domain, and (2) from synthetic image domain into the real image domain. We fine-tune the pose estimation network by constraining that to extract features with common domain knowledge between synthetic and real data.   
  
    \item Extensive experiments on the evaluation of each proposed component of FiDIP tested on real infant pose images, which  shows that our  method provides consistent performance improvement when applied on the existing state-of-the-art (SOTA) human pose estimation models with both complex and light-weight pose predictor backbones. This allows the implementation of FiDIP on embedded systems (e.g. baby monitors) for long-term as well as real-time pose monitoring of infants. 
\end{itemize}

\section{Related Work}
\label{sec:related}
\paragraph{Infant Pose Estimation.} For applications that require infant posture/motion analysis, the current approaches are dominantly based on (real-time or recorded) visual observation by the infant's pediatrician or the use of contact-based inertial sensors \cite{airaksinen2020automatic}. Meanwhile, there exist very few recent attempts initiated by the computer vision community to automatically perform pose estimation and tracking on videos taken from infants. In \cite{hesse2017body}, the authors estimated 3D body pose of infants in depth images for their motion analysis purpose, however they only evaluated their method on simple supine positions from limited number of subjects. They employed a pixel-wise body part classifier using random ferns to predict infant's 3D joints in order to automate  the task of motion analysis for identifying infantile motor disorders. In \cite{hesse2018learning}, the authors presented a statistical learning method called 3D skinned multi-infant linear (SMIL) body model using incomplete low quality RGB-D sequence of freely moving infants. The specific dataset they used is provided in \cite{hesse2018computer}, where users mapped real infant movements to the SMIL model with natural shapes and textures, and generated RGB and depth images with 2D and 3D joint positions. However, both of these works rely heavily on having access to the RGBD data sequence, which is difficult to obtain and hinder the use of these algorithms in regular webcam-based baby monitoring systems.

\paragraph{Synthetic Human Pose Data Generation.}
Synthesizing complicated articulated 3D models such as a human body has been drawing huge attention lately due to its extensive applications in studying human poses, gestures, and activities. Among benefits of synthesizing data is the possibility to automatically generate enough labeled data for supervised learning purposes, especially in small data domains. In \cite{liu2018semi}, the authors introduce a semi-supervised data augmentation approach that can synthesize large-scale labeled pose datasets using 3D graphical engines based on a physically-valid low dimensional pose descriptor. As introduced in \cite{rhodin2018unsupervised}, 3D human poses can be reconstructed by learning a geometry-aware body representation from multi-view images without annotations. Another research trend in synthesizing human pose images is simulating human figures by employing generative adversarial network (GAN) techniques. The authors in \cite{ma2017pose} present a two-stage pose-guided person generation network to integrate pose by feeding a reference image and a novel pose into a U-Net-like network to generate a coarse reposed person image, and refine image by training the U-Net-like generator in an adversarial way. In these works, however, neither the generated human avatars nor the reconstructed poses are able to accurately adapt to the infant style. Additionally, these GAN-based approaches of synthetic human figures do not have the capabilities of simulating complicated poses regularly taken by infants. 

Based on the above-mentioned challenges in achieving a robust infant pose estimation model and the shortcomings of the prior arts, 
we address the problem by contributing: (1) a hybrid real and synthetic infant pose dataset (SyRIP) that benefits from both realistic poses/appearances as well as synthesizing augmentation, (2) a fine-tuned domain-adapted infant pose (FiDIP) approach which shows consistent improvement over conventional fine-tuning evaluated on  several SOTA backbones (see \figref{visualized}). 

\visualized

\section{SyRIP: Synthetic/Real Infant Pose Dataset for Pose Data Augmentation}
\label{sec:dataset}
As stated earlier, there is a shortage of labeled infant pose dataset, and despite recent efforts in developing them, a versatile dataset with different and complex poses to train a deep network on is yet to be built. The only publicly-available infant image dataset is MINI-RGBD dataset \cite{hesse2018computer}, which provides only 12 synthetic infant models with continuous pose sequences. However, beside having simple poses, MINI-RGBD sequential feature leads to a small variation in the poses between adjacent frames and the poses of whole dataset are mainly repeated. In \figref{Distri}(a), we show the distribution of body poses of MINI-RGBD dataset and observe that poses in this dataset are simple and lack variations. Both its simplicity and being exclusively synthetic would cause the pose estimation models trained on MINI-RGBD to not generalize well to the real-world infant images. 

However, collecting and fully annotating a large-scale real infant pose dataset comparable to the size of adult pose datasets is very challenging, due to the privacy concerns that have also limited the number of samples in the web. To address this data limitation, we present a hybrid dataset forming strategy by mixing both real and synthetic infant pose data to form our SyRIP dataset, which exploits not only the real infant pose and appearance information, but also the data augmentation flexibility through using a set of synthetic infant models.

\Distri

\subsection{Real Infant Pose Data Gathering}
\label{sec:real}
Due to the difficulties in controlling infant movements as well as critical privacy concerns in collecting images from someone's child, access to infant images with wide variety of poses is limited. For real portion of the SyRIP dataset, we look for publicly available yet scattered real infant images from sources such as \textit{YouTube} and \textit{Google Images}. We follow the convention on working with the publicly available images and each image’s URL link is provided in our dataset. The biggest benefit of this collection method is that the diversity of the infant poses is guaranteed to the greatest extent. We choose infants (newborn to one year old) in various poses and many different backgrounds. 

We manually query YouTube and download more than 40 videos with different infants, and then split each video sequence to pick about 12 frames containing different poses. Finally, about 500 images including more than 50 infants with different poses from those frames are collected. We also select about 200 high-resolution images containing more than 90 infants from the Google Images. Compared to images taken from the YouTube videos, images from Google Images with higher resolution can be used to improve the quality of the whole dataset. The pose distribution of the real part of SyRIP dataset is shown in \figref{Distri}(b). Obviously, these poses are more diverse than those in the MINI-RGBD dataset.

\subsection{Cross Domain Inspired Synthetic Data Augmentation}
\label{sec:synthetic}
200 real images are far too small to train a deep neural network  and even not enough to fine-tune a pose estimation model with deep structure. Simulation seems to be a valid way to augment the dataset \cite{varol2017learning}, however it comes out to be challenging for infants as there are neither many infant 3D scans available to augment their appearance data, nor any infant motion captured movements to augment their pose data. 
Here we propose a cross domain inspired synthetic augmentation approach for infant pose data simulation. The pipeline of synthetic augmentation is illustrated in \figref{SynPipeline}.

We employ the SMIL model \cite{hesse2018computer}  for our synthetic data generation, which has $N = 6890$ vertices and $K = 23$ joints, and can be parameterized by the pose coefficients $\theta \in \mathbb{R}^{3(K+1)}$, where $K+1$ stands for body joints and one more joint (i.e. pelvis, is the root of the kinematic tree) for global rotation, and the shape coefficients $\beta \in \mathbb{R}^{20}$, representing the proportions of the individual's height, length, fat, thin, and head-to-body ratio. The infant mesh is then given as $M(\beta, \theta)$ and a synthetic image $I_{syn}$ is generated through the imaging process $\mathcal{I}$ as: 
\begin{equation}
\setlength{\abovedisplayskip}{8pt}
\setlength{\belowdisplayskip}{8pt}
    I_{syn} = \mathcal{I}(M(\beta, \theta), C(d,f), Tx, Bg),
\end{equation}
where $C$ represents the camera parameters depending on the camera principal point $d$ and focal length $f$. $Tx$ stands for the texture and $Bg$ stands for the background.  
We augment the camera parameter with random position with a fixed focal length. For the background, we pick 600 scenarios approximately related to infant indoor and outdoor activities from LSUN dataset \cite{yu15lsun}. Unfortunately, SMIL provides only limited appearances and simple pose parameters. There are neither known infant motion capture data nor extra infant appearances  for SMIL model. To augment these parameters, we employ references from neighboring domains as following:

\parskip=10pt\noindent \textbf{Inspiration from the real infant pose domain}: Although we do not have a versatile 3D infant pose data for pose augmentation, however these infant specific poses are actually reflected in the real infant images scrapped from web though in 2D.  We first employ the SMPLify-x approach \cite{pavlakos2019expressive} to lift these 2D poses into the SMIL pose $\theta$ by minimizing the cost function as:
\begin{equation}
\label{eqn:error}
\setlength{\abovedisplayskip}{8pt}
\setlength{\belowdisplayskip}{8pt}
L = L_J(\beta, \theta; C, j_{2D}) + \lambda_\theta L_\theta(\theta) + \lambda_\beta L_\beta(\beta) + \lambda_\alpha
L_\alpha(\theta),
\end{equation}
where $C$ is intrinsic camera parameters,  $\lambda_\theta$, $\lambda_\beta$, and $\lambda_\alpha$ are weights for specific loss terms, as described  in \cite{pavlakos2019expressive}. \eqnref{error} is the sum of four loss terms: (1) $L_J$ a joint-based data term, which is the distance between groundtruth 2D joints $j_{2D}$ and the 2D projection of the corresponding posed 3D joints of SMIL for each joint, (2) $L_\theta$ defined as a mixture of Gaussians pose prior learnt from 37,000 adult poses \cite{pavlakos2019expressive}, (3) a shape penalty $L_\beta$, which is the Mahalanobis distance between the shape prior of SMIL and the shape parameters being optimized, and (4) a pose prior penalizing elbows and knees $L_\alpha$. Therefore, we can augment the synthetic infant pose and shape via learned parameter from the real images.

\parskip=10pt\noindent\textbf{Inspiration from the adult appearance domain}: 
Even different in size,  infants still share the similar kinematic structure as adults and most adult poses are kinematically  compatible with infants. With the same topology as the adult template SMPL \cite{loper2015smpl}, many existing adult scans' textures can be transferred directly into the infant models. 
A valid concern could be that with adult's  textures the model will be no infant-like. However, we argue that with limited data, these borrowed variations can prevent overfitting and improve the model robustness. Therefore, we not only utilize the 12 infant textures (naked only with diaper) provided by MINI-RGBD dataset, but also augment appearance with adult textures from 478 male and 452 female clothing images coming from synthetic humans for real (SURREAL) dataset \cite{varol2017learning}.

\parskip=0pt During synthesizing, we manually filter out the unnatural/invalid generated infant bodies and add random noise term into augmented pose data to further increase its variance.   We visualize the pose distribution of the SyRIP synthetic subset, in \figref{Distri}(c), to make sure that the poses in our synthetic dataset has enough variations.

Finally, a new infant pose dataset, synthetic and real infant pose (SyRIP), is built up including both real and synthetic images that display infants in various positions, and utilize it to train pose estimation models with our proposed FiDIP method. Our process includes a training part consists of 200 real and 1000 synthetic infant images, and a test part with 500 real infant images, all with fully 2D body joints annotated efficiently by utilizing an AI-human co-labeling toolbox (AH-CoLT) \cite{huang2019ah} which is our previous work. Infants in these images have many different poses, like crawling, lying, sitting, and so on. The pose distributions of MINI-RGBD, SyRIP real part and SyRIP synthetic part are shown in \figref{Distri}, in which SyRIP dataset shows noticeably more pose variations compared the existing infant pose dataset, MINI-RGBD. 
\SynPipeline

\section{FiDIP: Fine-tuned Domain-adapted Infant Pose Estimation}
\label{sec:net}
Our FiDIP approach makes use of an initial pose estimation model trained on the abundant adult pose data, then fine-tunes that model on the augmented dataset, which consists of a small amount of real infant pose data and a series of pose-diverse synthetic infant images. For the augmented dataset, a domain adaptation method is proposed to align features of synthetic infant data with the real-world infant images. As the number of images in our dataset is limited, we only update a few layers of that network to fine-tune that for infant pose estimation rather than re-training the whole adult pose estimation network.

\subsection{FiDIP Framework} 
\label{sec:arch}
Our FiDIP framework is integrable with any existing encoder-decoder pose model. As illustrated in \figref{NetPipeline}, a pose estimation model with feature extractor as its encoder and pose estimator as its decoder could apply FiDIP by introducing a domain classification head. The entire model can be treated as two sub-network: pose estimation network and domain confusion network. Pose estimation network could be any pose model including SimpleBaseline \cite{xiao2018simple}, DarkPose \cite{zhang2020distribution}, and Hourglass \cite{newell2016stacked}. The domain confusion network, which is composed of a feature extractor shared with the pose estimation component and a domain classifier, is added to enforce the images in the real or synthetic domain being mapped into a same feature space after feature extraction. Domain classifier is designed to be a binary classifier with only three fully connected layers to distinguish whether the input feature belongs to a real or synthetic image. In particular, the domain confusion network assists pose estimation network during training. At test time, only the pure pose model (pose estimation network) works independently.

\subsection{Network Training} 
\label{sec:train}
The FiDIP training procedure consists of initialization session and a formal training session where the domain classifier and feature extractor are trained in a circular way. 

\noindent\textbf{Model initialization:} The pose estimation component of FiDIP network is already pre-trained on adult pose images from COCO dataset \cite{lin2014microsoft}. Since our training strategy is based on the use of fine-tuning as a means for transfer learning, to avoid unbalanced components' updating during fine-tuning, the domain classifier part of our domain confusion sub-network also needs to be pre-trained on both real and synthetic data from adult humans in advance. This combination dataset includes real adult images from the validation part of COCO dataset and some part of SURREAL dataset \cite{varol2017learning}. During pre-training, the feature extractor part stays frozen, and only weights for domain classifier will be initialized. The following stages are done after this initialization.

\noindent\textbf{Formal training session:} In this session, for each iteration the network is updated in a circular way with two stages:
\hspace*{0.25cm}

\textbf{Stage I.} In this stage, we lock pose estimation sub-network and fine-tune the domain classifier of domain confusion sub-network based on the current performance of feature extractor using infant real and synthetic pose data. The objective of this stage is to obtain a domain classifier for predicting whether the features are from a synthetic infant image or real one. Since the pose estimation network is locked and only domain classifier is to be optimized, the optimization objective in this stage is the loss of domain classifier $L_D$, which is calculated by the binary cross entropy:
\vspace{-.2cm}
\begin{equation}
\label{eqn:lossD}
L_{D} = -\frac{1}{N} \sum_{i=1}^{N} d_{i} \cdot \log f(s_{i}) + (1 - d_{i}) \cdot \log (1-f(s_{i})),
\end{equation}
where $s_{i}$ is the score of $i$th feature belonging to synthetic domain, $d_{i}$ is the corresponding groundtruth, $f(\cdot)$ represents the Sigmoid function, and $N$ is the batch size.

\textbf{Stage II.} The pose estimation network is to be fine-tuned with locked domain classifier in this stage. We try to refine the feature extractor to not only affect the pose predictor but also confuse the domain classifier. We leverage the domain classifier updated at stage I to promote the feature extractor to retain the ability to extract keypoints' information during the fine-tuning process, but also to ignore the differences between the real domain and the synthetic domain. An adversarial training method, which is proposed in \cite{ganin2015unsupervised}, is utilized to pushing features from synthetic images and real images into a common domain. A gradient reversal layer (GRL) is introduced to minimize the pose loss ($L_P$).

To train our domain classifier in a balance way,
we propose a balancing strategy by increasing the weight of real data during training. The $L_P$ loss, which measures the mean squared error between predicted heatmap/coordinates $\hat{y}_{i}$ and targeted heatmap/coordinates $y_{i}$ for each keypoint $i$, is: 

\vspace{-.2cm}
\begin{equation}
\label{eqn:lossP_2}
L_{P} = \frac{1}{N} \sum_{i=1}^{N} S(I_i)(\hat{y}_{i} - y_{i})^{2},
\end{equation}
where $S(I_i)$ is the scaling factor in the domain indicator $I_i$. 
It simultaneously maximizes the domain loss ($L_D$), so that the features representing both synthetic and real domains become similar. The optimization objective is:
\vspace{-.1cm}
\begin{equation}
\label{eqn:lossT}
L(\theta_{f}, \theta_{y}, \theta_{d}) = L_{P}(\theta_{f}, \theta_{y})-\lambda L_{D}(\theta_{f}, \theta_{d}),
\end{equation}
where $\lambda$ controls the trade-off between the two losses that shape the features during fine-tuning. $\theta_f$, $\theta_y$, and $\theta_d$ represent parameters of feature extractor, pose predictor, and domain classifier, respectively. 

\section{Experimental Evaluation}
\label{sec:experiments}
Our solution towards a robust infant pose estimation comes from two main contributions: (1) SyRIP dataset and (2) FiDIP approach. In this section,  we evaluate each component specifically. 

\subsection{Datasets}
Two infant datasets are employed in our evaluation, one is the only public  infant pose dataset, called MINI-RGBD, and the other one is our SyRIP dataset. We also employ 1904 samples from COCO val2017 \cite{lin2014microsoft} and 2000 random images from SURREAL \cite{varol2017learning} during our pre-training stage. 

MINI-RGBD dataset \cite{hesse2018computer} has 12 synthetic infant models with their continuous pose sequences. SyRIP dataset includes 700 real infant images with representative poses via manually selection and 1000 synthesized infants. For a reliable evaluation, we keep a large portion, 500 images, of real infant data as test set which we call Test500 for our common test. The rest 200 real with the synthetic infant data is used at the training set. 
One observation in our study is that, many pre-trained models are able to accurately estimate infant poses that are similar to adult poses, however not for infant unique poses such as bend legs over the chest. 
To evaluate a model's performance over specific infant poses, we collect a challenging subset with 100 complex yet typical infant poses from Test500 which we call Test100. Some samples are provided in \textit{Supplementary Materials}.

It is clear that the number of images in our test set is much smaller compared to the datasets used in other human pose estimation studies. Indeed, due to the aforementioned limitations caused by privacy, security, and other objective conditions, obtaining sufficient amount of infant pose images (that can publicly get access to) is an ongoing challenge, which makes our application a clear example in ``Small Data'' domain. We make up for the lack of data scale by enriching the poses, characters, and scenes in our SyRIP dataset.

\subsection{Implementation Details}
We employ several SOTA pose estimation structures with varying complexity as our backbone network, including the ResNet-50 of SimpleBaseline (SimpleBaseline-50) \cite{xiao2018simple}, HRNet-W48 of DarkPose \cite{zhang2020distribution} and MobileNetV2 \cite{sandler2018mobilenetv2} to reflect the general effect of our FiDIP framework.   
We add our domain classifier which has 3 fully connected layers on top of the backbone output features. For DarkPose, we choose the highest resolution branch. 
During training, we employ Adam optimizer with learning rate of 0.001. The batch size and epoch for initialization session are 128 and 1, respectively. While, for formal training session, there are 100 epochs and 64 images in a batch. During the Stage II, we set GRL parameter $\lambda$ as 0.0005, and freeze the first three layers (Res1, Res2, and Res3) of the feature extractor in our detailed ablation study. As for evaluation metric, we employ mean average precision (mAP) \cite{lin2014microsoft} over 10 thresholds of the object keypoint similarity (OKS), which is the distance between predicted keypoints and ground truth keypoints normalized by the scale of the person.

\subsection{Evaluation Over SyRIP} 
We gauge the SyRIP quality by specifically evaluating the effect of its synthetic data as well as its real and synthetic hybrid data.  In a straight forward way, we compare identical models fine-tuned on SyRIP or MINI-RGBD datasets to compare their performances as shown in \tabref{syripabl}.

\syripabl

To evaluate SyRIP quality, we employ three SOTA pose estimation models to fine-tune on MINI-RGBD, SyRIP-syn (synthetic portion only) and SyRIP whole set and compare their performance as shown in \tabref{syripabl}. From the result, we can see that with limited synthesized appearances and limited poses, the model tuned on MINI-RGBD is easily overfitted with even lower performance than the original model.  In comparison, in our CDIA approach by extensively learning from neighboring domains, the data variation is increased and even with our synthetic infant data alone, `SyRIP-syn', and without any adaptation, the model performance is still improved. 
Additional real infant data as in full SyRIP set, further increase the performance that indicates the benefit of our hybrid strategy. 
All these improvements are observed on all tested models with varying computational complexities.

\subsection{Evaluation over FiDIP}
For the infant pose estimation problem, two hypotheses may be assumed: (1) 2D human pose estimation models trained on the large-scale public datasets will be universally effective on different subjects, including infants.  (2) If not, they can be fine-tuned with a few samples from the target domain to achieve high performance.  In this section, we evaluate these hypotheses by comparing: (a) FiDIP with SOTA pre-trained models; (b) FiDIP ablation study; (c) FiDIP with conventional fine-tuning approach. 
For fair comparison,  all models are trained on SyRIP if needed and the performance advantage purely comes from the approaches. 

\comp

\paragraph{Comparison with the SOTA general purpose pose estimation models} 
We compare FiDIP model with a ResNet50 backbone \cite{xiao2018simple} with pre-trained SOTA approaches as shown in \tabref{comp}. Obviously, most models are well-performed on SyRIP Test500, which indicates infant and adult share many common poses. However for infant-specific poses in Test100, their performance drops dramatically as these poses are rarely seen among adults. 
In comparison, our approach (FiDIP) shows noticeably better results in both Test100 and Test500. We can see that pre-trained SOTA human pose models are not universally effective and infant pose estimation can be improved significantly via our approach. 

We also provide qualitative visualizations of our SimpleBaseline+FiDIP model on SyRIP test dataset compared to the Faster R-CNN, DarkPose, and SimpleBaseline models performance in \figref{visualized}. Simple poses, such as the examples in the 1st row of \figref{visualized}, are predicted accurately by almost all of the SOTA models. However, in infant's daily activities, their poses are often varied and more complex, especially in their lower body. DarkPose model based on ResNet-50 with $128\times96$ input size (2nd column) and Faster R-CNN model based on ResNet-50 (3rd column) trained on the adult datasets, show obvious inaccuracies in localizing the position of infant's legs and feet. Even SimpleBaseline based on ResNet-50 and DarkPose based on HRNet \cite{DBLP:journals/corr/abs-1908-10357} models with $384\times288$ input size are unable to keep high performance of infant lower body estimation. SimpleBaseline+FiDIP has much greater chance of inferring keypoints correctly for infant pose images than other models as shown in \figref{visualized}. 

\ablation
\paragraph{Ablation study}
\tabref{ablation} investigates the performance of alternative choices of FiDIP on SimpleBaseline-50 (SimpleBaseline based on ResNet-50) model, where method \textbf{n} is our well-performed FiDIP model as reported in \tabref{comp}.

\fDistri

\textit{Domain Adaptation.} We explore whether the domain adaptation method we implement can effectively overcome the difference between feature spaces of the real (R) domain and synthetic (S) domain in our SyRIP training dataset, so we test on 700 real images and 1000 synthetic images from the whole SyRIP dataset (1200 training + 500 testing) for easier observation. Methods that contain domain adaptation show higher AP than other method without domain adaption. t-SNE \cite{maaten2008visualizing} is used to visualize the distributions of extracted features for original SimpleBaseline-50, method \textbf{i}, and method \textbf{n} in \figref{fDistri}. Obviously, the FiDIP method embedded with domain adaptation component can align the feature distribution better than other networks.

\textit{Update Layers.} Freezing weights of the first few layers of the pre-trained network is a common practice when fine-tuning network with an insufficient amount of training data. The first few layers are responsible to capture universal features like curves and edges, so we fix them to enforce our network to focus on learning dataset-specific features in the subsequent layers at Stage II. We explore the effect of updating different numbers of last few layers of network on the performance of the trained model. In \tabref{ablation}, for method \textbf{m} and \textbf{n}, the ResNet 4th and 5th blocks of our feature extractor (ResNet-50) are updated, while the first four ResNet blocks are fixed and only the weights of last one block are updated in method \textbf{k} and \textbf{l}. We observe that method \textbf{m}, \textbf{n} perform much better than the other two.

\paragraph{Comparison with Direct Fine-Tuning}
A classical approach for transfer learning is a straightforward fine-tuning. Here, we employ three SOTA backbones for our pose estimation models with varying complexity, Pose-MobileNet, DarkPose, and SimpleBaseline, and compare FiDIP version and fine-tuned version head to head with result shown in \tabref{mobile}. To achieve pose estimation goal on backbone MobileNetV2, the Pose-MobileNet is built by adding a pose regressor as a decoder behind MobileNetV2. We initially train it on COCO Train2017 to get a pre-trained model and then fine-tune or apply FiDIP method to Pose-MobileNet on SyRIP dataset. 

\mobile

\section{Conclusion}
In this paper, we present a solution towards robust infant pose estimation, which includes an infant dataset SyRIP with hybrid  synthetic and real data and a FiDIP strategy to transfer learn from existing adult models and datasets. Our FiDIP framework consists of a pose estimation sub-network to leverage transfer learning from a pre-trained adult pose estimation network and a domain confusion sub-network for adapting the model to both real infant and synthetic infant datasets. With identical network structure, we compared their performance when trained on our SyRIP dataset and on the only other publicly available infant pose dataset MINI-RGBD respectively to show the benefit of dataset forming strategy with high data scarcity challenge. With identical dataset, we compared FiDIP approach with the fine-tuning approach to show the advantages of FiDIP across multiple pose estimation  models with various complexities. 

\vspace{-.06in}

{\small
\bibliographystyle{ieee}
\bibliography{paper}
}

\supp
\end{document}